\documentclass[letterpaper]{article}

\usepackage{aaai2027}
\usepackage[hyphens]{url}
\usepackage{graphicx}
\usepackage{natbib}
\usepackage{caption}
\usepackage{booktabs}
\usepackage{algorithm}

\usepackage{microtype}
\usepackage{subcaption}
\usepackage{amsmath}
\usepackage{amssymb}
\usepackage{mathtools}
\usepackage{amsthm}
\usepackage{pifont}

\theoremstyle{plain}

\theoremstyle{definition}

\theoremstyle{remark}

\newif\ifappendixattached
\appendixattachedtrue

\title{Rethinking Federated Graph Foundation Models: A Graph-Language Alignment-Based Approach}

\author{
    Yinlin Zhu\textsuperscript{\rm 1},
    Di Wu\textsuperscript{\rm 1}\corresponding,
    Xianzhi Zhang\textsuperscript{\rm 1},
    Yuming Ai\textsuperscript{\rm 2},\\
    Xunkai Li\textsuperscript{\rm 2},
    Miao Hu\textsuperscript{\rm 1},
    Guocong Quan\textsuperscript{\rm 1}
}

\affiliations{

    Sun Yat-sen University, Guangzhou, China\textsuperscript{\rm 1}\\
    Beijing Institute of Technology, Beijing, China\textsuperscript{\rm 2}\\

    \{zhuylin27,\ zhangxzh9\}@mail2.sysu.edu.cn, 
    yumingai@bit.edu.cn, cs.xunkai.li@gmail.com,\\
    \{wudi27,\ humiao5,\ quangc\}@mail.sysu.edu.cn
}

\begin{document}

\maketitle

\begin{abstract}
  Recent studies of federated graph foundational models (FedGFMs) break the idealized and untenable assumption of having centralized data storage to train graph foundation models, and accommodate the reality of distributed, privacy-restricted data silos.
  Despite their simplicity and intuition, existing studies that project aligned generalizable knowledge onto a discrete token space via vector-quantized backbones suffer from irreversible knowledge loss during the quantization process.
  In this context, we argue that reconciling the semantic-structural orthogonality and integrity between pre-trained language models (PLMs) and graph neural networks (GNNs) is paramount for developing effective FedGFMs while simultaneously mitigating the severe data heterogeneity and communication constraints inherent in distributed, resource-limited environments. To address these issues, we propose \textbf{FedGALA} (\underline{\textbf{Fed}}erated \underline{\textbf{G}}raph \underline{\textbf{A}}nd \underline{\textbf{L}}anguage \underline{\textbf{A}}lignment), a framework that resolves graph-based semantic-structural orthogonality and integrity in federated settings by employing unsupervised contrastive learning to align GNNs and frozen PLMs within a continuous embedding space, thereby capturing robust, transferable general knowledge.
  Subsequently, FedGALA leverages a communication-efficient prompt tuning mechanism to steer these pre-aligned encoders and frozen PLMs, facilitating effective adaptation to diverse downstream tasks while circumventing the prohibitive overhead of full-parameter fine-tuning.
  The comprehensive experiments validate that FedGALA outperforms all competitive baselines across multi-domain datasets on multiple tasks with up to 14.37\% performance improvement. 
\end{abstract}

\section{Introduction}
The success of pre-trained language models (PLMs) has catalyzed the rise of generalizable foundation models~\citep{Langgfm,xia2024gfm_anygraph,gfm_gfse} across diverse downstream tasks~\citep{cai2021link_prediction2,zhang2019graph_classification1,gfm_swapgt}. This paradigm naturally extends to text-attributed graphs (TAGs) through the emergence of graph foundation models (GFMs). 
While GFMs leverage the synergy between textual semantics and complex topology, their reliance on centralized corpora is often untenable given the fragmented, privacy-restricted nature of real-world data silos~\citep{fu2022fgl_survey_1, zhang2021fgl_survey_2, li2024openfgl}. 
Consequently, federated graph learning (FGL) has emerged as the imperative solution for collaborative training without raw data transfer~\citep{zhang2021fedsage,zhu2024fedtad}, yet the development of robust federated graph foundation models (FedGFMs) remains a critical and underexplored frontier. 
Constructing effective FedGFMs requires addressing three fundamental dimensions:

\ding{182} \textbf{Semantic-Structural Orthogonality:} The disconnect between sequential language and non-Euclidean topology necessitates a dual-encoder architecture that is difficult to synchronize in decentralized settings. 
PLMs lack structural inductive biases, while textual neighborhood summaries are limited by narrow receptive fields. 
In FedGFMs, topological heterogeneity~\citep{li2024fedgta,li2024adafgl} across silos impedes the alignment of semantic and structural spaces without direct data access. 
This dichotomy complicates knowledge generalization, leading to cross-domain entanglement.

\ding{183} \textbf{Cross-domain Knowledge Entanglement:} FedGFM generalizability is limited by fragmented local datasets. Severe non-IID distributions, manifested as semantic and structural divergence across clients, hinder the learning of unified, transferable representations. The divergence exacerbated by disparate domains causes conflicting local motifs that trigger overfitting and obstruct the acquisition of transferable knowledge, preventing robust global model convergence. 

\ding{184} \textbf{Communication Overhead:} Bandwidth bottlenecks in federated settings make exchanging massive PLM parameters infeasible. 
Developing practical FedGFMs requires bandwidth-efficient paradigms that adapt frozen models to downstream tasks while maintaining semantic-structural alignment under strict physical network constraints.

Pioneering frameworks like FedGFM+~\citep{zhu2025fedgfm} and FedBook~\citep{fedbook} utilize vector-quantized (VQ) methods~\citep{gfm_gft} as backbones.
However, these approaches reveal critical deficiencies when evaluated against the aforementioned fundamental challenges:

\textbf{Regarding Challenge 1}: Projecting high-dimensional semantic spaces onto discrete tokens induces irreversible quantization errors. This discretization loses the subtle interplay between textual semantics and complex topologies, which severely limits structural-semantic fidelity.

\textbf{Regarding Challenge 2}: VQ methods trade representational completeness for statistical generalizability by forcing diverse local distributions onto a constrained global codebook, leading to representation homogenization and neglecting nuanced client-specific long-tail knowledge.

\textbf{Regarding Challenge 3}: Transmitting discrete tokens creates a suboptimal equilibrium that sacrifices model convergence quality for communication efficiency.

These limitations call for a new FedGFM training paradigm. To this end, we propose \textbf{FedGALA} (\underline{\textbf{Fed}}erated \underline{\textbf{G}}raph \underline{\textbf{A}}nd \underline{\textbf{L}}anguage \underline{\textbf{A}}lignment), which leverages unsupervised contrastive learning to bridge the semantic-structural gap in continuous latent spaces. By prioritizing representational fidelity through continuous alignment, FedGALA enables the framework to harness PLMs' generalized knowledge for enhanced cross-domain transferability and robust few-shot performance, establishing a superior foundation for the two-stage training pipeline: \underline{\textit{Federated Pre-training Phase}} for cross-domain graph encoding and \underline{\textit{Prompt-based Fine-tuning Phase}} for task adaptation. Specifically, in order to resolve \textbf{Challenge 1}, FedGALA aligns the frozen PLM textual features with the topology-oriented GNN features; similarly, to address \textbf{Challenge 2}, it disentangles the graph encoder into local semantic and shared structural components, while also using semantic perturbation and a history-aware ensemble to stabilize domain-invariant global updates; and finally, to overcome \textbf{Challenge 3}, it aggregates only the lightweight structural encoders and employs prompt tuning to efficiently adapt the frozen backbones.

\textbf{Our Contributions:} \ding{182} \textbf{New Perspective.} We pinpoint representation loss in quantized FedGFMs and propose continuous structural-semantic alignment as a robust training alternative. \ding{183} \textbf{New Framework.} FedGALA bridges semantic-structural gaps through federated contrastive pre-training followed by communication-efficient local prompt fine-tuning. \ding{184} \textbf{Superior Performance.} Comprehensive evaluations confirm that FedGALA surpasses 22 SOTA baselines with significant performance gains reaching 14.37\%.

\section{Preliminaries}

\subsection{Problem Formalization}
The FedGFM framework comprises $K$ distributed clients where each client $k$ possesses a private subgraph $G_k$ and textual attributes $\mathcal{T}_k$, aiming to collaboratively train a graph encoder $f_{\theta}$ while the text encoder $g_{\psi}$ remains frozen. Pre-training phase establishes a modality-invariant space by minimizing a global contrastive loss $\mathcal{L}_{pre}$ that uses weights $\alpha_k$ to aggregate across-domain structural knowledge.
\begin{equation}
    \min_{\theta} \mathcal{L}_{pre} = \sum_{k=1}^K \alpha_k \mathcal{L}_{local}(G_k, \mathcal{T}_k; \theta, \psi),
\end{equation}
In the fine-tuning phase, backbone models, such as $\theta_{frozen}$, $\psi_{frozen}$ remain frozen to shift optimization toward a personalized prompt pool $\mathcal{P} = \{\phi_1, \dots, \phi_M\}$. Each client identifies the best-fitting prompt index $m_k^*$ to minimize the task-specific loss $\mathcal{L}_{task}$ for local data distributions $D_k$.
\begin{equation}
    \min_{\phi_{m_k^*}} \mathcal{L}_{task}(D_k; \theta_{frozen}, \psi_{frozen}, \phi_{m_k^*}).
\end{equation}
\subsection{Graph Foundation Model (GFM)}
GFMs~\citep{gfm_samgpt, Langgfm} utilize hybrid architectures where frozen PLMs extract semantic features and trainable GNNs encode topological structures for multi-modal expertise. 
Existing research employs self-supervised objectives like contrastive learning~\citep{GraphCL, GCC} or cross-modal alignment~\citep{TEA_GLM, GraphAdapter, gfm_gofa} to unify textual and structural knowledge before downstream adaptation.
Benchmark system~\citep{GFM_Benchmark} has recently been proposed to evaluate centralized GFMs systematically on their transferability across multiple domains. 
Centralized GFMs are unrealistic: single centers lack the massive data scale required for transferability, and privacy laws prevent sharing raw data, making a collaborative paradigm essential.

\subsection{Federated Graph Foundation Model (FedGFM)}
FedGFMs integrate the multi-modal expertise of GFMs into a privacy-preserving collaborative training framework. FedGFM+~\citep{zhu2025fedgfm} utilizes anchor-based initialization and domain-sensitive prompt pools for personalization. 
FedBook~\citep{fedbook} employs a unified federated codebook to model intra-domain and inter-domain knowledge.
Both approaches using discrete vector quantization suffer from irreversible knowledge loss. To preserve representational fidelity, a redesign is necessary to align GNNs with frozen PLMs via self-supervised contrastive learning in a continuous latent space.
FedGALA addresses this through a bandwidth-efficient framework that achieves multi-modal alignment for robust structural knowledge transfer.

\section{Methods}
\begin{figure*}[t]
  \centering
  \includegraphics[width=0.9\textwidth]{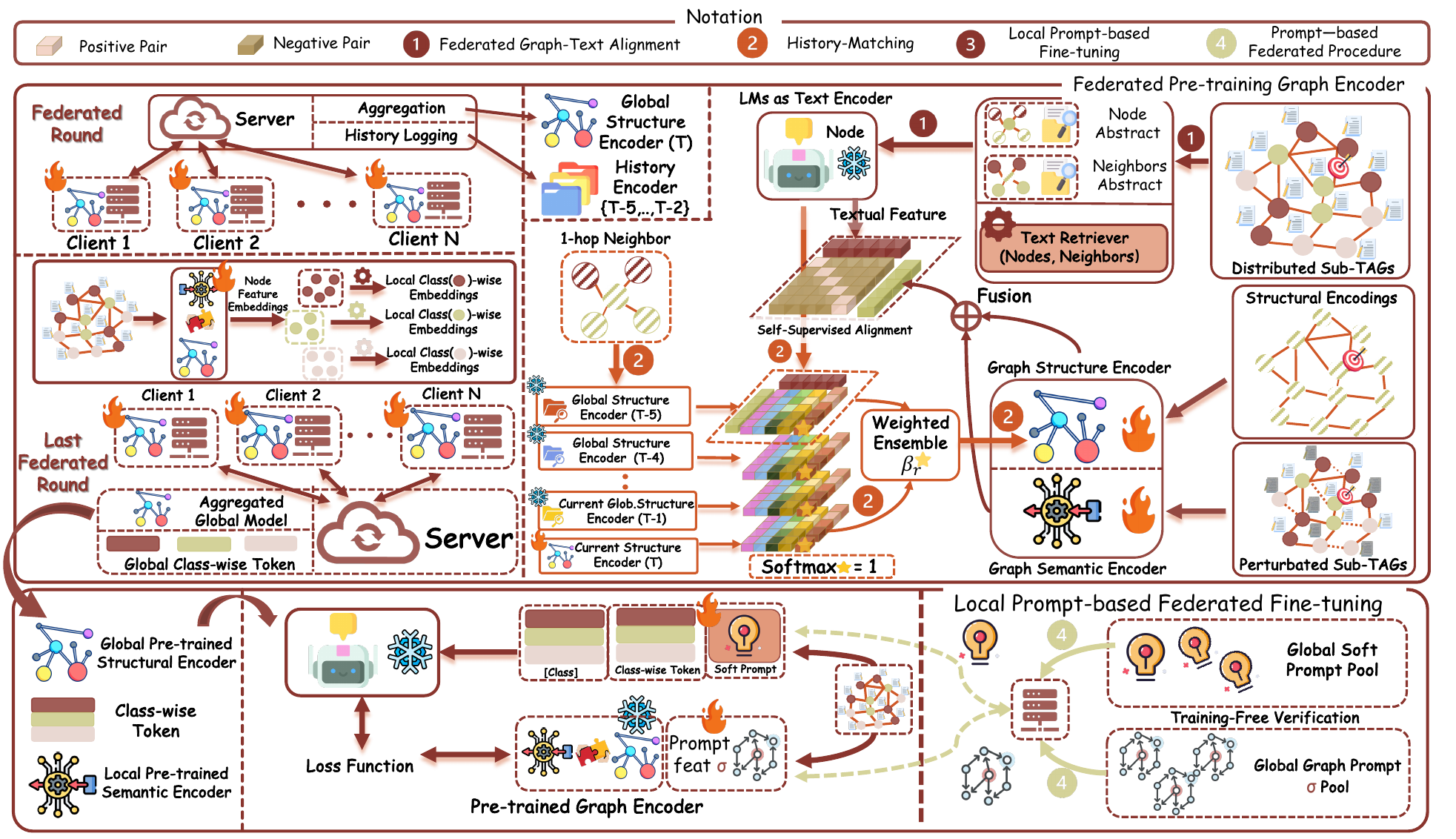}
  \caption{\textbf{FedGALA framework} contains two training phases:
  \ding{182} \textbf{Federated Pre-training Graph Encoder:} aligning frozen PLMs with partitioned graph encoders (structural and semantic) via contrastive alignment and history-matching to derive global structural parameters, class-wise tokens, and local semantic encoders.
  \ding{183} \textbf{Local Prompt-based Federated Fine-tuning:} adapting the frozen PLMs to various downstream tasks, and the trained prompts are consolidated via group-aware prompt aggregation.
  }
  \label{fig: framework}

\end{figure*}

FedGALA is a training pipeline that integrates GNNs with a prompt-based tuning mechanism to adapt frozen PLMs across heterogeneous local datasets. 
The framework is designed in two phases to maximize global knowledge acquisition while ensuring local task-specialized performance.  

\ding{182} \textbf{Pre-training Phase} (Sec.\ref{subsec: pre-training}) facilitates collaborative learning across diverse domains to produce a graph encoder with transferable global structural and local semantic knowledge. 
During this phase, clients also generate class-wise embeddings that the server aggregates into global class-wise tokens to provide semantic anchors for providing additional guidance to frozen PLMs in the second phase.

\ding{183} \textbf{Local Fine-tuning Phase} (Sec.\ref{subsec: Fine-tuning}) begins with pre-trained components sent to clients, which remain frozen.
Soft prompts and global class-wise tokens guide the frozen PLMs toward specific downstream tasks based on local data adaptability. 
This dual-phase design captures universal structural motifs unattainable through isolated training and provides critical guidance to prevent overfitting on limited local datasets.
Each client receives the most effective soft and graph prompts from global pools and proceeds to a group-aware prompt aggregation mechanism to update these pools.  
The overall visual framework is shown in Fig.~\ref{fig: framework}, and the complete training procedure is summarized in the technical appendix.
\subsection{Federated Pre-training Graph Encoder}
\label{subsec: pre-training}
The pre-training phase develops a graph encoder with robust cross-domain transferability while addressing non-IID data distribution challenges in the FL setting.
Standard aggregation of full parameters often causes knowledge conflicts and hinders convergence due to the entanglement of local node features and global structural patterns.
To mitigate this, FedGALA decouples graph knowledge by distinguishing between domain-specific semantic features and invariant structural patterns. 
This strategy grounds semantic knowledge locally while collaboratively training a global structural encoder that possesses universal topological motifs~\citep{xie2021gcfl,tan2023fgl_fedstar} across diverse datasets. 
The subsequent sections detail this two-step procedure.

\subsubsection{Federated Graph-Text Alignment}
FedGALA utilizes self-supervised contrastive learning to align the latent representations of the graph encoder $G_{\theta}$ and the text encoder $T_{\psi}$. 
By minimizing a symmetric contrastive loss, the model pulls positive graph-text pairs $(\mathbf{z}_i^G, \mathbf{z}_i^T)$ closer in the embedding space while separating non-corresponding pairs.
The aim is to extract invariant, transferable global structural knowledge through collaboration. 
Two key mechanisms achieve this intention:

\ding{182} \textbf{Context-Aware Semantic Summarization:} To leverage textual knowledge effectively, we adopt a node summary strategy where a structural-semantic summary $\mathcal{T}_i$ is constructed by concatenating target node descriptions with aggregated descriptions of their one-hop neighborhood $\mathcal{N}_i$. 
This anchoring provides the text encoder with a local field of view and ensures the semantic representation $\mathbf{z}_i^T = g_{\psi}(\mathcal{T}_i)$ remains grounded in graph topology.

\ding{183} \textbf{Decoupled Structural-Semantic Encoding:} To maximize the acquisition of transferable structural knowledge while preventing semantic interference, the graph encoder is partitioned into two specialized components:
\underline{\textit{Structural}} \underline{\textit{Encoder}} processes positional encodings (PE) and the raw adjacency matrix $\mathbf{A}_k$ to focus on topological connectivity. \underline{\textit{Semantic Encoder}} processes node features $\mathbf{X}_k$ and the adjacency matrix $\mathbf{A}_k$ using a feature disturbance mechanism to minimize reliance on idiosyncratic local semantics. 
The above encoding results are fused to obtain the $\mathbf{z}_i^G$.
This decoupling forces the objective to rely on invariant structural knowledge through the local contrastive loss $\mathcal{L}_{local}$ defined for $N$ samples, which is calculated as follows:
\begin{equation}
    s_{i,j} = \frac{(\mathbf{z}_i^G)^\top \mathbf{z}_j^T}{\|\mathbf{z}_i^G\| \|\mathbf{z}_j^T\|} \cdot \exp(\tau),
\end{equation}
\begin{equation}
\label{eq: loss function}
\begin{split}
    \mathcal{L}_{local} = -\frac{1}{2N} \sum_{i=1}^N \Bigg[ & \log \frac{\exp(s_{i,i})}{\sum_{j=1}^N \exp(s_{i,j})} \\
    & + \log \frac{\exp(s_{i,i})}{\sum_{j=1}^N \exp(s_{j,i})} \Bigg],
\end{split}
\end{equation}
where $\tau$ is a learnable temperature coefficient. 
During the federated phase, only structural encoder parameters $\theta_{str}$ are transmitted to reduce bandwidth and prevent semantic drift.
The server employs topology-aware aggregation weights $\alpha_k$ based on the average degree $\bar{d}_k$ of each client local graph:
\begin{equation}
    \bar{d}_k = \frac{1}{|\mathcal{V}_k|} \sum_{v \in \mathcal{V}_k} \text{deg}(v), \qquad \alpha_k = \frac{\bar{d}_k}{\sum_{j=1}^K \bar{d}_j}.
\end{equation}
The global structural parameters $\theta_{global}^{(t+1)}$ are computed as:
\begin{equation}
\label{eq: pre-training aggregation}
    \theta_{global}^{(t+1)} = \sum_{k=1}^K \alpha_k \theta_k^{(t)}.
\end{equation}
This design prioritizes clients with denser connectivity to facilitate the learning of universal structural representations.

\subsubsection{Federated History-Matching}
This module aims to stabilize structural knowledge acquisition by mitigating client drift and preventing the catastrophic forgetting of learned topological motifs. 
Unlike standard FGL training that only preserves the latest global updates, the server of FedGALA maintains a temporal log of the $R=5$ most recent global structural encoders $\{\theta_{\text{his}}^{(r)}\}_{r=1}^R$ and broadcasts this pool to all clients. 
Each client then identifies the historical parameters that best align with its local data manifold through a three-step matching probe:

\ding{182} \textbf{Structural Forward Pass:} Local structural inputs are processed by the current global model and the historical versions to generate candidate embeddings $\hat{\mathbf{Z}}_{\text{his}, r}^G$.

\ding{183} \textbf{Semantic Alignment:} The text encoder generates a stable semantic anchor $\hat{\mathbf{Z}}^T$ by leveraging aggregated summaries of the local 1-hop neighborhood information.

\ding{184} \textbf{Similarity Weighting:} Each client computes cosine similarities between the candidate structural outputs and the semantic anchor to determine the corresponding weights $\beta_r$ via a softmax function.

Based on the acquired weight $\beta_r$ for each logged global model, each client performs a weighted aggregation of historical parameters to optimize the local initialization for the next training round. 
The similarity scores and the resulting learnable parameter fusion with the locally-preserved semantic encoder $\theta_{\text{local}}$ are defined as:
\begin{equation}
    \mathcal{S}_r = \mathbb{E}_{i \sim \mathcal{B}} \left[ \frac{\langle \phi(G_i; \theta_{his}^{(r)}), \psi(\mathcal{T}_i) \rangle}{\|\phi(G_i; \theta_{his}^{(r)})\|_2 \|\psi(\mathcal{T}_i)\|_2} \right],
\end{equation}
\begin{equation}
    \boldsymbol{\beta} = \text{Softmax} \left( \sum_{i \in \mathcal{D}_k} \left[ \mathcal{S}_1^{(i)}, \dots, \mathcal{S}_R^{(i)} \right]^\top \right),
\end{equation}
\begin{equation}
\label{eq: hist-matching fusion}
    \theta_{graph} \leftarrow \theta_{local} + \sum_{r=1}^R \beta_r \theta_{his}^{(r)}.
\end{equation}
By integrating historical global knowledge that aligns with local topological patterns, this module ensures a consistent and robust structural learning trajectory.

\subsubsection{Federated Pre-trained Class-wise Token}
After the pre-training (i.e., start fine-tuning), FedGALA generates weighted class prototypes to provide global class-wise guidance.
Such a design delivers a comprehensive label distribution that transcends the restricted, often skewed, local data silos of individual clients. 
This global guidance is essential to prevent the frozen text encoder from overfitting to local semantic distributions during the subsequent task adaptation.
To ensure the uploaded local class-wise embeddings are a reliable representation of local knowledge for global aggregation, this module introduces two metrics: 

\ding{182} \textbf{Knowledge Strength ($s_i^{\text{str}}$):} By utilizing the maximum logit probability as a proxy for predictive confidence, this metric ensures that the global representation is primarily shaped by high-certainty nodes. 
It selectively amplifies robust signals while suppressing noise in heterogeneous local graph datasets.

\ding{183} \textbf{Knowledge Clarity ($s_i^{\text{clr}}$):} This metric addresses the over-smoothing phenomenon by penalizing nodes with excessive similarity to neighbors. By prioritizing nodes with distinct decision boundaries, the prototype preserves high-frequency structural details often lost in collaborative training.

The local class-wise prototype $\mathbf{P}_c$ is computed via weighted aggregation of embeddings $\mathbf{z}_i^G$ within each class $c$:
\begin{equation}
\begin{split}
    \omega_i = {} & \underbrace{\max_{c \in \mathcal{C}} P(y_i = c | \mathbf{z}_i^G)}_{\text{Knowledge Strength } s_i^{\text{str}}} \\
    & + \underbrace{\Big( 1 - \frac{1}{k} \sum_{j \in \mathcal{N}_k(i)} \frac{\mathbf{z}_i^\top \mathbf{z}_j}{\|\mathbf{z}_i\| \|\mathbf{z}_j\|} \Big)}_{\text{Knowledge Clarity } s_i^{\text{clr}}},
\end{split}
\end{equation}
\begin{equation}
\label{eq: class-wise token aggregation}
    \mathbf{P}_c = \frac{\sum_{i \in \mathcal{V}_{k,c}} \omega_i \cdot \mathbf{z}_i^G}{\sum_{i \in \mathcal{V}_{k,c}} \omega_i}.
\end{equation}
These local prototypes are transmitted to the server, where a clustering-based aggregation strategy resolves semantic conflicts across domains by grouping similar prototypes. 
\subsection{Local Prompt-based Federated Fine-tuning}
\label{subsec: Fine-tuning}
The second phase aims to adapt the pre-trained foundation to specific downstream tasks while navigating a computational and communication-wise efficient strategy. 
Traditional fine-tuning is hindered by the massive parameter scale of PLMs and the risk of catastrophic interference caused by non-IID data distributions. 
To address these, FedGALA keeps the pre-trained graph and text encoders frozen to preserve the structural-semantic alignment established during the pre-training phase. Instead of costly backbone updates, we introduce a personalized prompt tuning mechanism that optimizes a lightweight pool of learnable prompts.
This strategy enables clients to accommodate local data nuances and diverse tasks without incurring prohibitive computational overhead or corrupting global foundational knowledge.

\subsubsection{Prompt Pool Initialization and Selection}
FedGALA utilizes a dual-channel prompt pool mechanism to balance global knowledge transfer with local task adaptation while mitigating the costs of full-parameter fine-tuning. The server constructs learnable global textual and graph prompt pools, denoted as $\mathcal{P}_T = \{\phi_1^T, \dots, \phi_M^T\}$ and $\mathcal{P}_G = \{\phi_1^G, \dots, \phi_M^G\}$. 
To accelerate convergence, initialized prompts are guided by class-wise tokens from the pre-training phase, anchoring them in a pre-aligned multimodal latent space. 
Each client $k$ evaluates the entire pool against its local data $D_{k, val}$ and frozen backbone $\theta_{frozen}$ to identify optimal indices $m_{k, T}^*$ and $m_{k, G}^*$ :
\begin{equation}
\label{eq: prompt selection}
    m_{k, \text{type}}^* = \underset{m \in \{1, \dots, M\}}{\arg \max} \; \mathcal{E} \left( \mathcal{D}_{k, val}; \Theta_{\text{frozen}}, \phi_m^{\text{type}} \right),
\end{equation}
for $\text{type} \in \{T, G\}$. This process serves as a distributional filter that each client begins fine-tuning with prompts most relevant to its domain.

\subsubsection{Local Prompt-based Fine-tuning}
Following selection, clients adapt multi-modal knowledge to specific tasks while preserving the structural and semantic alignment established during pre-training. 
Adhering to parameter-efficient fine-tuning (PEFT) principles, the graph encoder ($G_{\theta}$) and text encoder ($T_{\psi}$) remain frozen, which prevents catastrophic forgetting of learned universal motifs and ensures computational feasibility by restricting optimization to lightweight prompt parameters ($\phi_{m_k^*}^T, \phi_{m_k^*}^G$). 
Using a task-specific loss $\mathcal{L}_{\text{task}}$, the local update rule is:
\begin{equation}
\label{eq:2nd phase loss function =}
\begin{split}
    \phi_{m_k^*}^{\text{type}} \leftarrow {} \phi_{m_k^*}^{\text{type}} - \eta \nabla_{\phi^{\text{type}}} \mathcal{L}_{\text{task}} \big( \mathcal{D}_k; \Theta_{\text{frozen}}, \{ \phi_{m_k^*}^{T}, \phi_{m_k^*}^{G} \} \big),
\end{split}
\end{equation}
for $\text{type} \in \{T, G\}$, where $\eta$ is the learning rate. This enables prompts to bridge the foundation model toward local target tasks without compromising its core cross-domain capabilities.
\subsubsection{Group-aware Prompt Aggregation}

The final stage performs targeted updates based on the functional specialization of the prompt pool. The server organizes updates into client subsets $\mathcal{S}_m$ based on selected indices $m$ and conducts independent weighted aggregation:
\begin{equation}
\label{eq: prompt aggregation}
    \phi_m^{(t+1)} = 
    \begin{cases} 
        \sum_{k \in \mathcal{S}_m} \frac{n_k}{N_m} \phi_k^{(t)}, & \text{if } \mathcal{S}_m \neq \emptyset; \\[1.5ex]
        \phi_m^{(t)}, & \text{otherwise},
    \end{cases}
\end{equation}
where $n_k$ is the sample count and $N_m$ is the total volume. 
Non-selected prompts remain stationary ($\phi_m^{(t+1)} = \phi_m^{(t)}$) to prevent the dilution of specialized knowledge by unrelated domains. 
It fosters a multi-expert system where prompts represent specific latent domains, enabling high performance across diverse tasks without gradient interference.

Due to space constraints, we detail the computational and communication overhead for FedGALA relative to FedBook and FedGFM+ within Appendix~\ref{appendix: complexity analysis} for reference.

\section{Experiment}
We evaluate FedGALA to answer the following questions.
\textbf{Q1}: Does FedGALA outperform existing FedGFMs and conventional FL/FGL baselines across downstream tasks? (Sec.\ref{subsec: Comparison})
\textbf{Q2}: Can FedGALA outperform other FedGFMs in 2-shot learning settings? (Sec.\ref{subsec: 2-shot learning})
\textbf{Q3}: What contributes to the success of FedGALA? (Sec.\ref{subsec: Ablation})
\textbf{Q4}: How do the pre-defined second-phase hyperparameters impact FedGALA? (Sec.\ref{subsec: sensitivity analysis})
\textbf{Q5}: How efficient is FedGALA compared to existing FedGFMs? (Sec.\ref{subsec: Efficiency_Analysis})

\begin{table*}[t]
    \centering
    \small
    \begin{tabular}{l|cccc|cc|cc}
        \toprule
        \textbf{Method}
        & \textbf{Cora}
        & \textbf{PubMed}
        & \textbf{OGB-arxiv}
        & \textbf{WikiCS}
        & \textbf{FB15K237}
        & \textbf{WN18RR}
        & \textbf{HIV}
        & \textbf{PCBA} \\
        \midrule
        Linear
        & 72.37$_{\pm 0.33}$
        & 82.20$_{\pm 0.04}$
        & 67.76$_{\pm 0.15}$
        & 72.27$_{\pm 0.04}$
        & 67.77$_{\pm 0.36}$
        & 79.83$_{\pm 0.30}$
        & 60.46$_{\pm 0.28}$
        & 57.75$_{\pm 0.18}$ \\
        GCN
        & 75.22$_{\pm 0.19}$
        & 82.07$_{\pm 0.27}$
        & \underline{70.68}$_{\pm 0.07}$
        & \underline{75.54}$_{\pm 0.33}$
        & 66.95$_{\pm 0.38}$
        & 79.74$_{\pm 0.14}$
        & 60.30$_{\pm 0.37}$
        & 63.15$_{\pm 0.22}$ \\
        GAT
        & \underline{76.13}$_{\pm 0.16}$
        & 82.75$_{\pm 0.09}$
        & 68.02$_{\pm 0.23}$
        & 74.40$_{\pm 0.22}$
        & 68.59$_{\pm 0.19}$
        & \underline{81.35}$_{\pm 0.31}$
        & 60.12$_{\pm 0.20}$
        & 66.41$_{\pm 0.23}$ \\
        GraphSAGE
        & 74.61$_{\pm 0.20}$
        & \underline{83.53}$_{\pm 0.44}$
        & 68.31$_{\pm 0.22}$
        & 74.87$_{\pm 0.31}$
        & \underline{69.94}$_{\pm 0.32}$
        & 80.33$_{\pm 0.06}$
        & 61.02$_{\pm 0.10}$
        & 66.50$_{\pm 0.08}$ \\
        GIN
        & 74.65$_{\pm 0.27}$
        & 81.92$_{\pm 0.23}$
        & 68.28$_{\pm 0.44}$
        & 74.62$_{\pm 0.22}$
        & 68.37$_{\pm 0.40}$
        & 79.94$_{\pm 0.39}$
        & \underline{61.38}$_{\pm 0.15}$
        & \underline{67.31}$_{\pm 0.24}$ \\
        \midrule
        FedAvg
        & 75.76$_{\pm 0.19}$
        & 82.84$_{\pm 0.26}$
        & 70.35$_{\pm 0.08}$
        & 75.31$_{\pm 0.07}$
        & 68.43$_{\pm 0.29}$
        & 80.07$_{\pm 0.12}$
        & 63.95$_{\pm 0.05}$
        & 68.62$_{\pm 0.42}$ \\
        MOON
        & 76.43$_{\pm 0.26}$
        & 85.44$_{\pm 0.41}$
        & 71.19$_{\pm 0.08}$
        & 76.37$_{\pm 0.04}$
        & 68.44$_{\pm 3.48}$
        & 80.62$_{\pm 0.42}$
        & 65.96$_{\pm 0.01}$
        & 69.42$_{\pm 0.10}$ \\
        FedSage+
        & 76.32$_{\pm 0.16}$
        & \underline{85.78}$_{\pm 0.11}$
        & 71.98$_{\pm 0.12}$
        & 76.22$_{\pm 0.17}$
        & \underline{69.95}$_{\pm 0.29}$
        & \underline{81.26}$_{\pm 0.26}$
        & \textit{N/A}
        & \textit{N/A} \\
        Fed-PUB
        & 76.33$_{\pm 0.35}$
        & 85.01$_{\pm 0.22}$
        & 72.09$_{\pm 0.20}$
        & \underline{77.39}$_{\pm 0.17}$
        & 69.32$_{\pm 0.29}$
        & 80.65$_{\pm 0.30}$
        & \textit{N/A}
        & \textit{N/A} \\
        FedGTA
        & 76.43$_{\pm 0.26}$
        & 83.16$_{\pm 0.26}$
        & 72.13$_{\pm 0.08}$
        & 76.68$_{\pm 0.28}$
        & \textit{N/A}
        & \textit{N/A}
        & \textit{N/A}
        & \textit{N/A} \\
        FedTAD
        & \underline{76.55}$_{\pm 0.23}$
        & 84.52$_{\pm 0.32}$
        & 72.35$_{\pm 0.16}$
        & 75.56$_{\pm 0.12}$
        & \textit{N/A}
        & \textit{N/A}
        & \textit{N/A}
        & \textit{N/A} \\
        FGSSL
        & 74.41$_{\pm 0.26}$
        & 83.73$_{\pm 0.44}$
        & \underline{72.62}$_{\pm 0.10}$
        & 75.90$_{\pm 0.21}$
        & \textit{N/A}
        & \textit{N/A}
        & \textit{N/A}
        & \textit{N/A} \\
        FGGP
        & 74.73$_{\pm 0.17}$
        & 83.66$_{\pm 0.07}$
        & 71.62$_{\pm 0.10}$
        & 76.41$_{\pm 0.48}$
        & \textit{N/A}
        & \textit{N/A}
        & \textit{N/A}
        & \textit{N/A} \\
        GCFL+
        & \textit{N/A}
        & \textit{N/A}
        & \textit{N/A}
        & \textit{N/A}
        & \textit{N/A}
        & \textit{N/A}
        & \underline{67.34}$_{\pm 0.24}$
        & 71.27$_{\pm 0.16}$ \\
        FedStar
        & \textit{N/A}
        & \textit{N/A}
        & \textit{N/A}
        & \textit{N/A}
        & \textit{N/A}
        & \textit{N/A}
        & 66.20$_{\pm 0.28}$
        & \underline{71.75}$_{\pm 0.33}$ \\
        \midrule
        OFA$^*$
        & 75.37$_{\pm 0.57}$
        & 84.65$_{\pm 0.14}$
        & 70.10$_{\pm 0.15}$
        & 77.96$_{\pm 0.22}$
        & 70.24$_{\pm 0.18}$
        & 81.85$_{\pm 0.19}$
        & \underline{69.54}$_{\pm 0.20}$
        & 72.30$_{\pm 0.39}$ \\
        GFT$^*$
        & 76.66$_{\pm 0.65}$
        & 84.79$_{\pm 0.34}$
        & \underline{73.58}$_{\pm 0.15}$
        & \underline{78.37}$_{\pm 0.60}$
        & 71.77$_{\pm 0.37}$
        & 81.20$_{\pm 0.31}$
        & 67.57$_{\pm 0.59}$
        & 71.52$_{\pm 0.18}$ \\
        UniGraph$^*$
        & 77.95$_{\pm 0.24}$
        & 85.25$_{\pm 0.36}$
        & 71.25$_{\pm 0.25}$
        & 75.08$_{\pm 0.11}$
        & 72.24$_{\pm 0.22}$
        & 82.33$_{\pm 0.17}$
        & 68.40$_{\pm 0.40}$
        & 72.16$_{\pm 0.10}$ \\
        GQT$^*$
        & \underline{79.20}$_{\pm 0.37}$
        & \underline{85.79}$_{\pm 0.30}$
        & 71.69$_{\pm 0.43}$
        & 76.61$_{\pm 0.45}$
        & \underline{72.81}$_{\pm 0.19}$
        & \underline{83.73}$_{\pm 0.42}$
        & 68.32$_{\pm 0.68}$
        & 71.93$_{\pm 0.31}$ \\
        GraphCLIP$^*$
        & 78.56$_{\pm 0.27}$
        & 85.65$_{\pm 0.33}$
        & 72.86$_{\pm 0.41}$
        & 77.08$_{\pm 1.11}$
        & 72.56$_{\pm 0.09}$
        & 83.58$_{\pm 0.60}$
        & 66.75$_{\pm 3.60}$
        & \underline{72.58}$_{\pm 0.26}$ \\
        \midrule
        FedGFM+
        & 80.59$_{\pm 0.34}$
        & 87.17$_{\pm 0.05}$
        & 74.51$_{\pm 0.21}$
        & 78.69$_{\pm 0.16}$
        & 73.23$_{\pm 0.21}$
        & 84.31$_{\pm 0.63}$
        & 69.70$_{\pm 0.32}$
        & 74.81$_{\pm 0.31}$ \\
        FedBook
        & \underline{81.32}$_{\pm 0.23}$
        & \underline{88.19}$_{\pm 0.25}$
        & \underline{76.21}$_{\pm 0.07}$
        & \underline{79.87}$_{\pm 0.28}$
        & \underline{74.33}$_{\pm 0.24}$
        & \underline{86.21}$_{\pm 0.35}$
        & \underline{70.10}$_{\pm 0.06}$
        & \underline{75.41}$_{\pm 0.15}$ \\
        \textbf{FedGALA (Ours)}
        & \textbf{84.60}$_{\pm 0.18}$
        & \textbf{92.48}$_{\pm 0.09}$
        & \textbf{80.10}$_{\pm 0.70}$
        & \textbf{83.60}$_{\pm 0.13}$
        & \textbf{77.85}$_{\pm 0.05}$
        & \textbf{88.33}$_{\pm 0.03}$
        & \textbf{73.30}$_{\pm 0.28}$
        & \textbf{77.75}$_{\pm 0.28}$ \\
        \bottomrule
    \end{tabular}
    \caption{Performance comparison between FedGALA and 22 baselines. The globally best and category-specific second-best results are marked in \textbf{bold} and \underline{underline}. `N/A' denotes the task inapplicability. `*' denotes federated adoption of GFMs.}
    \label{tab: performance comparison}
\end{table*}

\begin{table*}[t]
    \centering
    \small
    \begin{tabular}{l|cccccc}
        \toprule
        \textbf{Method}
        & \textbf{Cora}
        & \textbf{PubMed}
        & \textbf{OGB-arxiv}
        & \textbf{WikiCS}
        & \textbf{FB15K237}
        & \textbf{WN18RR} \\
        \midrule
        OFA$^*$
        & 50.23$_{\pm 0.34}$
        & 46.26$_{\pm 0.24}$
        & 17.11$_{\pm 0.36}$
        & 38.06$_{\pm 0.50}$
        & 19.44$_{\pm 0.42}$
        & 30.42$_{\pm 0.30}$ \\
        GFT$^*$
        & \underline{54.89}$_{\pm 0.15}$
        & 47.93$_{\pm 0.46}$
        & 18.76$_{\pm 0.35}$
        & \underline{40.48}$_{\pm 0.24}$
        & 18.99$_{\pm 0.29}$
        & 29.14$_{\pm 0.27}$ \\
        UniGraph$^*$
        & 49.80$_{\pm 0.23}$
        & 46.47$_{\pm 0.28}$
        & 18.74$_{\pm 0.16}$
        & 37.51$_{\pm 0.20}$
        & \underline{20.34}$_{\pm 0.38}$
        & 28.59$_{\pm 0.27}$ \\
        GQT$^*$
        & 53.11$_{\pm 0.48}$
        & 48.35$_{\pm 0.27}$
        & \underline{21.50}$_{\pm 0.31}$
        & 38.13$_{\pm 0.28}$
        & 20.16$_{\pm 0.17}$
        & \underline{30.66}$_{\pm 0.37}$ \\
        GraphCLIP$^*$
        & 51.11$_{\pm 0.26}$
        & \underline{50.31}$_{\pm 0.32}$
        & 20.84$_{\pm 0.17}$
        & 38.37$_{\pm 0.20}$
        & 19.74$_{\pm 0.33}$
        & 29.51$_{\pm 0.19}$ \\
        \midrule
        FedGFM+
        & 55.24$_{\pm 0.28}$
        & 52.10$_{\pm 0.30}$
        & 21.76$_{\pm 0.21}$
        & 42.27$_{\pm 0.23}$
        & 22.45$_{\pm 0.25}$
        & 32.32$_{\pm 0.22}$ \\
        FedBook
        & \underline{56.34}$_{\pm 0.35}$
        & \underline{55.23}$_{\pm 0.27}$
        & \underline{22.46}$_{\pm 0.43}$
        & \underline{45.05}$_{\pm 0.16}$
        & \underline{25.13}$_{\pm 0.27}$
        & \underline{33.29}$_{\pm 0.33}$ \\
        \textbf{FedGALA (Ours)}
        & \textbf{59.74}$_{\pm 0.16}$
        & \textbf{57.53}$_{\pm 0.29}$
        & \textbf{24.18}$_{\pm 0.30}$
        & \textbf{49.20}$_{\pm 0.21}$
        & \textbf{32.85}$_{\pm 0.25}$
        & \textbf{38.96}$_{\pm 0.26}$ \\
        \bottomrule
    \end{tabular}
    \caption{2-shot learning results for FedGALA, federated adaptations of centralized GFMs, FedGFM+, and FedBook.}
    \label{tab: few shot}
\end{table*}

\begin{table*}[t]
    \centering
    \small
    \begin{tabular}{l|cccc}
        \toprule
        \textbf{Method}
        & \textbf{OGB-arxiv}
        & \textbf{WikiCS}
        & \textbf{FB15K237}
        & \textbf{HIV} \\
        \midrule
        w/o History Matching
        & 77.28$_{\pm 0.18}$
        & 80.50$_{\pm 0.22}$
        & 74.49$_{\pm 0.17}$
        & 71.32$_{\pm 0.23}$ \\
        w/o Global Prompt Pool
        & 76.24$_{\pm 0.25}$
        & 78.61$_{\pm 0.19}$
        & 73.26$_{\pm 0.40}$
        & 68.32$_{\pm 0.22}$ \\
        \midrule
        \textbf{FedGALA}
        & \textbf{80.10}$_{\pm 0.70}$
        & \textbf{83.60}$_{\pm 0.13}$
        & \textbf{77.85}$_{\pm 0.05}$
        & \textbf{73.30}$_{\pm 0.28}$ \\
        \bottomrule
    \end{tabular}
    \caption{Ablation study: Evaluating history-matching and global prompt pool modules across diverse domains and tasks.}
    \label{tab: ablation study}
\end{table*}
\begin{figure}[t]
\centering
 \includegraphics[width=0.98\linewidth]{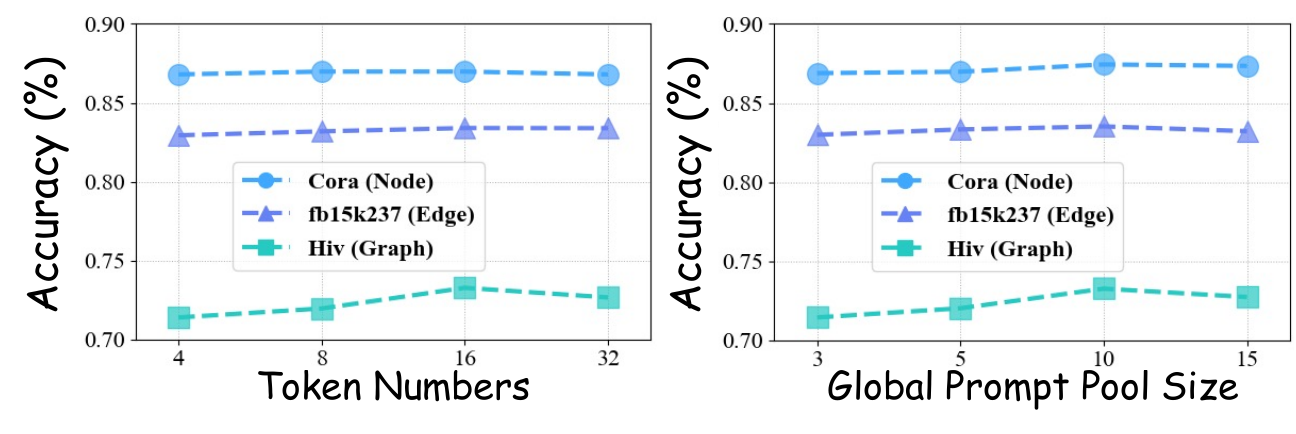}
	\caption{Sensitivity analysis of FedGALA.
    }
    \label{fig: sensitivity}
\end{figure}
\begin{figure}[t]
\centering
 \includegraphics[width=0.98\linewidth]{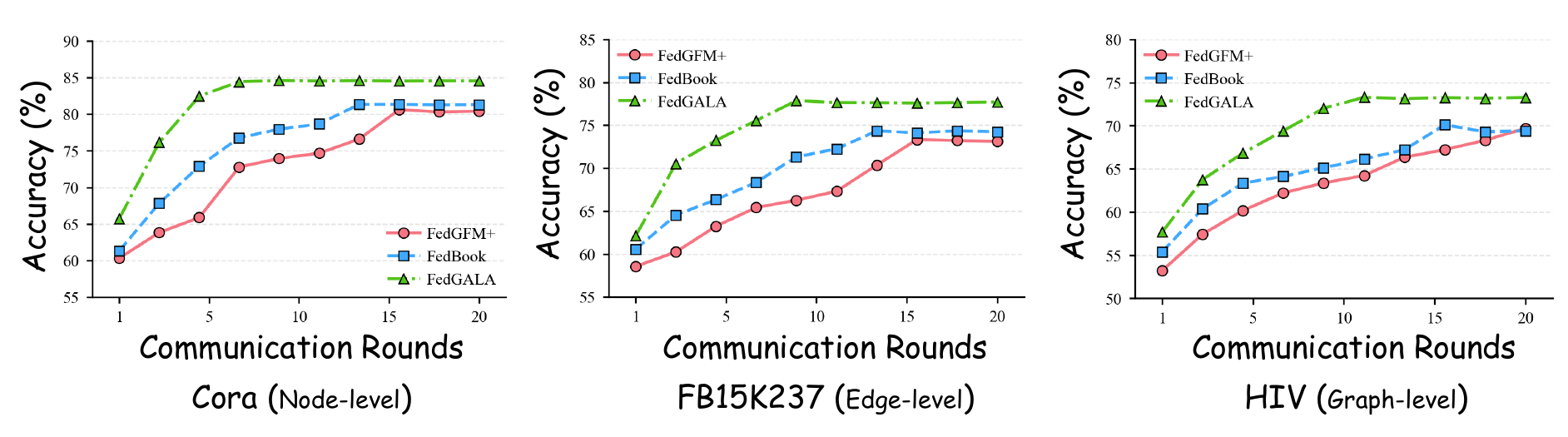}
	\caption{
    Convergence rates during the pre-training phase, where the proposed FedGALA consistently achieves faster convergence than all baseline methods.
    }
    \label{fig: convergence}
\end{figure}
\subsection{Experimental Setups}
\label{subsec: Experiment Setups}
This section introduces the experimental setup; due to space constraints, more details are provided in the appendices. 

\paragraph{Datasets.} We evaluate the proposed FedGALA on 8 benchmark graph datasets spanning 6 domains and 3 distinct tasks. Specifically, for \underline{\textit{Node Classification}}, we include Cora~\citep{Yang16cora}, PubMed, OGB-arxiv~\citep{hu2020ogb}, and WikiCS~\citep{mernyei2020wikics}. \underline{\textit{Edge Classification}} is performed on FB15K237~\citep{toutanova2015fb15k237} and WN18RR~\citep{dettmers2018dataset_wn18rr}, while \underline{\textit{Graph Classification}} utilizes the PCBA and HIV molecular datasets~\citep{wu2018dataset_moleculenet}. See Appendix~\ref{appendix: datasets} for more details.

\paragraph{Simulation Strategy.} 
To simulate the FedGFM scenario, each dataset is partitioned into three clients. Node and edge tasks utilize Metis~\citep{karypis1998metis} for topology-aware partitioning to preserve local structural integrity. Conversely, graph classification employs label-stratified random partitioning to ensure a balanced distribution of samples.

\paragraph{Baselines.} We compare FedGALA against 22 state-of-the-art baselines in four groups: (1) isolated supervised models, i.e., a linear classifier, GCN~\citep{kipf2016gcn}, GAT~\citep{velivckovic2017gat}, GraphSAGE~\citep{hamilton2017graphsage}, and GIN~\citep{xu2018gin}; (2) FL/FGL methods, i.e., FedAvg~\citep{mcmahan2017fedavg}, MOON~\citep{li2021moon}, FedSage+~\citep{zhang2021fedsage}, Fed-PUB~\citep{baek2022fedpub}, FedGTA~\citep{li2024fedgta}, FedTAD~\citep{zhu2024fedtad}, FGSSL~\citep{huangfgl_fgssl}, FGGP~\citep{wan2024fgl_fggp}, GCFL+~\citep{xie2021gcfl}, and FedStar~\citep{tan2023fgl_fedstar}; (3) federated adaptations of centralized GFMs, i.e., OFA~\citep{gfm_ofa}, GFT~\citep{gfm_gft}, UniGraph~\citep{gfm_unigraph}, GQT~\citep{gfm_gqt}, and GraphCLIP~\citep{gfm_graphclip}; (4) FedGFMs, i.e., FedGFM+~\citep{zhu2025fedgfm} and FedBook~\citep{fedbook}. Details of baseline descriptions can be found in Appendix~\ref{appendix: baseline}.

\subsection{Performance Comparison (Answer for Q1)}
\label{subsec: Comparison}

Table~\ref{tab: performance comparison} reports the comprehensive evaluation between FedGALA and baselines across various downstream tasks.
Overall, FedGALA achieves the best performance across all tasks and categories. 

\paragraph{Comparison to Isolated Supervised Models.}
Isolated supervised models highlight the limitations of fragmented local data, often outperforming naive federated strategies like FedAvg due to negative transfer and parameter entanglement. In contrast, FedGALA consistently achieves superior performance across all tasks by extracting universal structural knowledge through continuous multi-modal alignment. Specifically, it secures a 10.99\% gain on HIV graph classification and surpasses GraphSAGE on PubMed by 10.17\%. These results underscore the efficacy of FedGALA's collective intelligence in processing multi-modal data silos.

\paragraph{Comparison to FL/FGL Methods.} 
Traditional FL and FGL methods are architecturally limited to specific tasks, as evidenced by the `N/A' entries in Table~\ref{tab: performance comparison}. These approaches often fail to surpass isolated models because they cannot leverage multi-modal textual features, leading to suboptimal representations and knowledge entanglement.
FedGALA overcomes these constraints by utilizing textual and graph modalities through continuous alignment. Decoupling domain-specific semantics from invariant structural priors enhances generalization. Notably, FedGALA outperforms FedSage+ by 13.68\% on Cora, with improvements of 14.37\% and 10.56\% on FB15K237 and HIV, respectively.

\noindent \textbf{Comparison to Federated Adaptations of Centralized GFM Methods.}
We employ decentralized optimization with naive aggregation (e.g., FedAvg) for baselines in this category. While leveraging rich PLM textual features, these methods often suffer from knowledge entanglement and negative transfer because naive aggregation fails to balance intra-domain coherence with inter-domain diversity. In other words, they fail to navigate the multi-domain heterogeneity present in distributed data silos. FedGALA effectively mitigates these challenges through its specialized federated design, achieving up to 9.11\% gains over GraphCLIP in node, 3.56\% in link, and 10.91\% in graph-level tasks.

\noindent \textbf{Comparison to FedGFM Method.}
The results in Table~\ref{tab: performance comparison} demonstrate that FedGALA's continuous graph-text alignment significantly enhances representation quality compared to frameworks utilizing discrete VQ backbones. On average, FedGALA achieves performance gains of 5.41\% for node-level, 4.49\% for edge-level, and 4.17\% for graph-level tasks over existing FedGFMs. Notably, centralized GFMs under federated adaptation exhibit performance comparable to current FedGFMs, further validating the efficiency of a robust alignment mechanism in capturing transferable structural knowledge.

\subsection{Few-shot Learning Evaluation (Answer for Q2)}
\label{subsec: 2-shot learning}
To evaluate FedGALA's effectiveness in alleviating the cold-start problem for clients with scarce labeled data, we perform 2-shot learning experiments on node-level and edge-level tasks, omitting graph classification due to its multi-label complexity. We compare against centralized or federated GFMs, as they are viable for few-shot fine-tuning.
In Table~\ref{tab: few shot}, FedGALA achieves SOTA performance across all benchmarks, achieving 49.20\% accuracy on the 10-class WikiCS dataset, a 21.54\% improvement over GFT. Despite the complexity of OGB-arxiv's 40 classes, FedGALA outperforms the most competitive baseline by 7.66\%. Moreover, it exhibits a 10.87\% gain in edge classification. These results demonstrate that FedGALA's robustness stems from the high-fidelity, transferable knowledge acquired through its continuous alignment.

\subsection{Ablation Study (Answer for Q3)}
\label{subsec: Ablation}
Table~\ref{tab: ablation study} presents an ablation analysis focusing on two supplementary modules rather than phase removal, as the pre-training phase's efficacy is independently validated by the 2-shot results in Table~\ref{tab: few shot}. The federated history-matching module ensures structural knowledge reliability and prevents catastrophic forgetting by maintaining a temporal log of global encoders. Its removal causes declines of 3.10\% on WikiCS and 2.70\% on HIV. Similarly, the global prompt pool replaces naive equal-weight averaging with group-aware clustering to update global soft and graph prompts, preserving task specialization.
Without this mechanism, FedGALA suffers drops of 4.82\% on OGB-arxiv and 4.59\% on FB15K237, underscoring the necessity of mitigating the knowledge entanglement resulting from varying localized knowledge for efficient downstream adaptation.
\subsection{Sensitivity Analysis (Answer for Q4)}
\label{subsec: sensitivity analysis}
We conduct a sensitivity analysis to study the impact of token-related hyperparameters in the fine-tuning phase, specifically the token length and the number of soft prompts in the global pool. Fig.~\ref{fig: sensitivity} shows that FedGALA maintains stable performance across a wide range of settings on all three tasks. Optimal results are obtained with 16 soft prompt tokens and a pool of 10 global prompts, which we adopt as the recommended configuration. Effectiveness relies on balancing the pool size: large enough to capture multi-domain knowledge, yet small enough that each prompt receives dense updates from relevant clients; at 15 prompts, some remain under-updated, introducing underdeveloped priors that reduce training efficiency.

\subsection{Efficiency Analysis (Answer for Q5)}
\label{subsec: Efficiency_Analysis}
Fig.~\ref{fig: convergence} shows the pre-training convergence efficiency of FedGALA against existing FedGFM baselines. FedGALA achieves the fastest convergence across all tasks, stabilizing at round 6 for Cora, round 9 for FB15K237, and round 11 for HIV. In contrast, FedBook and FedGFM+ require approximately 15 to 20 rounds to converge. By transmitting only the lightweight structural encoders, FedGALA significantly reduces communication overhead compared to conventional federated settings, showing that it effectively mitigates knowledge conflicts and catastrophic forgetting.

\section{Conclusion}
Existing FedGFMs rely on discrete vector quantization, trading representation fidelity for computational efficiency. FedGALA introduces a new paradigm that emphasizes continuous semantic-structural alignment. Its two phase paradigm, federated pre training followed by communication efficient local prompt based fine tuning, captures transferable global structural knowledge while adapting frozen PLMs to diverse downstream tasks. Comprehensive evaluations demonstrate that FedGALA consistently outperforms state of the art baselines across multiple domains. Future work includes streamlining FedGALA for realistic deployment and extending
FedGFMs to multi modal graphs.

\bibliography{FedGALA}

\ifappendixattached
\newpage
\appendix

\section{Datasets Overview}
\label{appendix: datasets}
Eight datasets are being implemented in this work, and the statistical summary of them is exhibited in Table~\ref{tab: datasets}. 
The transition from standard graph neural network (GNN) benchmarks to text-attributed graphs (TAGs) marks a critical shift in the field, enabling the development of graph foundation models (GFMs). In standard versions of these datasets, node attributes are often pre-computed as shallow vectors (e.g., bag-of-words or word2vec) that lack the semantic nuance of the original text. The text-attributed versions provided here preserve or reconstruct raw textual descriptions, allowing large language models (LLMs) to serve as powerful encoders that unify disparate domains into a single semantic space.
The eight implemented datasets span citation networks, knowledge graphs, and molecular structures, categorized by their primary task level.

\ding{182} \textbf{Citation Network (Node-Level task)}: These datasets represent academic publications as nodes and citations as directed edges. In their TAG form, node attributes consist of paper titles and abstracts.

\textbf{Cora:} A standard citation network of 2,708 papers classified into 7 research areas. While the original version uses a binary bag-of-words vector, the TAG version utilizes the raw abstract text to capture deeper semantic relationships.

\textbf{PubMed}: A larger medical citation network (19,717 nodes) focusing on diabetes-related publications. The TAG version uses titles and abstracts to differentiate between 3 distinct classes.

\textbf{OGB-arxiv}: A massive, directed citation graph containing 169,343 CS papers. It uses pre-computed skip-gram embeddings, the TAG version allows encoders to process the original 128-dimensional textual metadata directly.

\textbf{WikiCS}: A hyperlink network based on 11,701 Computer Science-related Wikipedia entries. Nodes are enriched with entry names and article content, providing a dense textual corpus for 10-class classification.

\ding{183} \textbf{Knowledge Graphs (Edge-Level Tasks)}:
Knowledge graphs consist of entities (nodes) and relations (edges). TAG versions replace numerical IDs with natural language descriptions of entities and relations.

\textbf{FB15K237}: A subset of Freebase containing 14,541 entities and 237 relation types. The TAG implementation leverages entity names and descriptions to predict missing links.

\textbf{WN18RR}: A refined version of the WordNet knowledge graph (40,943 nodes). Text attributes provide linguistic definitions for 11 semantic relations (e.g., hypernym, member meronym), facilitating zero-shot link prediction.

\ding{184} \textbf{Molecular Graphs (Graph-Level Tasks)}:
In molecular datasets, each graph represents a single molecule, where nodes are atoms and edges are chemical bonds.

\textbf{PCBA}: A dataset comprising 437,929 molecules evaluated across 128 biological assays. In the TAG version, chemical properties and SMILES strings are transformed into textual descriptions of atomic and bond characteristics, allowing models to interpret the underlying chemistry of each molecule.

\textbf{HIV}: A dataset of 41,127 molecules screened for their ability to inhibit HIV replication. Text-based attributes allow GFMs to leverage pre-trained chemical-linguistic knowledge for binary classification of anti-HIV activity.

\section{Baseline Overview}
\label{appendix: baseline}
\begin{table*}[t]
\centering
\small
    \begin{tabular}{ccccccccc}
      \toprule
      \textbf{Dataset}  & \textbf{Domain} & \textbf{Task Level} & \textbf{\#Graphs} & \textbf{Avg. \#Nodes} & \textbf{Avg. \#Edges} & \textbf{\#Classes} \\ \midrule
      Cora    & Citation        & Node          & 1                  & 2,708                 & 10,556                & 7  \\
      PubMed   & Citation        & Node          & 1                  & 19,717                & 44,338                & 3     \\
      OGB-arxiv    & Citation        & Node          & 1                  & 169,343               & 1,166,243             & 40      \\
      WikiCS   & Hyperlink        & Node          & 1                  & 11,701                & 216,123               & 10    \\
      FB15K237 & Knowledge       & Link          & 1                  & 14,541                & 310,116               & 237       \\
      WN18RR   & Knowledge       & Link          & 1                  & 40,943                & 93,003                & 11      \\
      PCBA$^*$     & Molecule        & Graph         & 437,929            & 26.0                  & 28.1                  & 128\\
      HIV$^*$      & Molecule        & Graph         & 41,127             & 25.5                  & 27.5                  & 2  \\
      \bottomrule
    \end{tabular}
    \caption{Statistical summary of the experimental datasets, where multi-label classification datasets are denoted by the symbol `*'.}
\label{tab: datasets}
\end{table*}
\textbf{Isolated Supervised Learning Methods} are implemented using a dual-layer architecture with a hidden dimensionality of 64. For \textbf{FL/FGL Methods}, we employ task-specific backbones in instances where a custom architecture is not specified: GraphSAGE is utilized for node-level and edge-level classification, while GIN is adopted for graph-level tasks. Regarding the \textbf{Federated Adaptations of Centralized GFM Methods}, and \textbf{FedGFMs} we strictly maintain the architectural backbones as reported in their respective original studies.

\ding{182} \textbf{Isolated Supervised Learning.} To establish a performance baseline and assess the benefits of collaborative training, while also accounting for potential negative transfer, we evaluate standalone supervised models that are trained independently on each client. This category of local-only baselines operates without any federated communication and includes a standard linear layer as well as GCN~\citep{kipf2016gcn}, GAT~\citep{velivckovic2017gat}, GraphSAGE~\citep{hamilton2017graphsage}, and GIN~\citep{xu2018gin}.

\textbf{GCN}~\citep{kipf2016gcn} is a cornerstone of graph neural networks, utilizing spectral convolutions grounded in the normalized graph Laplacian to encode topology. By aggregating signals from adjacent nodes, it facilitates efficient node classification and remains a fundamental, computationally streamlined approach for graph representation learning.

\textbf{GAT}~\citep{velivckovic2017gat} adapts the self-attention paradigms prevalent in natural language processing to the graph domain. It introduces a dynamic weighting mechanism that enables nodes to selectively prioritize the most informative neighbors during aggregation, resulting in a highly adaptive learning process.

\textbf{GraphSAGE}~\citep{hamilton2017graphsage} addresses scalability by employing a sampling-driven message-passing framework. This architecture is particularly effective for inductive learning tasks, as it generalizes to previously unseen nodes by aggregating features from a sampled, fixed-size neighborhood. Its support for various learnable aggregation functions makes it well-suited for processing large-scale datasets.

\textbf{GIN}~\citep{xu2018gin} is engineered to maximize structural discriminative power, theoretically matching the Weisfeiler-Lehman graph isomorphism test in its capacity to distinguish complex graph topologies. Due to its emphasis on preserving structural integrity, GIN is frequently the preferred choice for graph-level representation tasks.
\ding{183} \textbf{FGL Approaches.} We benchmark our work against a diverse selection of FL and FGL baselines. This includes fundamental FL algorithms originally designed for computer vision tasks (FedAvg~\citep{mcmahan2017fedavg}, MOON~\citep{li2021moon}), as well as specialized subgraph-level FGL frameworks (FedSage+~\citep{zhang2021fedsage}, Fed-PUB~\citep{baek2022fedpub}, FedGTA~\citep{li2024fedgta}, FedTAD~\citep{zhu2024fedtad}, FGSSL~\citep{huangfgl_fgssl}, FGGP~\citep{wan2024fgl_fggp}) and graph-level FGL methods (GCFL~\citep{xie2021gcfl} and FedStar~\citep{tan2023fgl_fedstar}). The specific characteristics of these baselines are outlined below:

\textbf{FedAvg}~\citep{mcmahan2017fedavg} is a foundational FL protocol that enables private, decentralized model optimization. The process involves a central server distributing a global model to clients for local training; the resulting local parameters are then aggregated by the server to refine the global model for the subsequent iteration.

\textbf{MOON}~\citep{li2021moon} is a prominent FL method that employs model-level contrastive learning to synchronize local and global representations. It was developed to counteract performance degradation caused by non-IID data distributions across different clients.

\textbf{FedSage+}~\citep{zhang2021fedsage} combines the GraphSAGE~\citep{hamilton2017graphsage} architecture with the FedAvg~\citep{mcmahan2017fedavg} framework to facilitate FGL over local subgraphs. To enhance robustness against cross-client missing edges, it incorporates a neighbor generation mechanism that completes the local graph structure.

\textbf{Fed-PUB}~\citep{baek2022fedpub} offers a personalized subgraph-level FGL approach that functions without a shared global model. By utilizing functional embeddings from a random graph to calculate inter-client similarity, the server performs weighted aggregation. It further employs client-specific sparse masks to tailor model updates to the local subgraph.

\textbf{FedGTA}~\citep{li2024fedgta} scales graph learning within a federated context by having clients encode both node attributes and topological data. Clients compute local smoothing confidence and mixed feature moments, which the server then uses as weights to aggregate personalized models.

\textbf{FedTAD}~\citep{zhu2024fedtad} is a subgraph-centric method that utilizes topology-aware embeddings to assess the reliability of class-specific knowledge. This assessment allows the server to conduct data-free knowledge distillation, effectively migrating reliable insights from clients into the global architecture.

\textbf{FGSSL}~\citep{huangfgl_fgssl} mitigates the issue of client drift by harmonizing node-level semantics with global structural patterns. It uses contrastive objectives to align similar classes while isolating disparate ones, distilling relational knowledge from the global level into local client models.

\textbf{FGGP}~\citep{wan2024fgl_fggp} partitions the global model into a two-tiered structure connected by prototypes. At the decision level, class prototypes are used instead of standard classifiers to improve separation; at the representation level, contrastive learning is used to infuse prototypes with global knowledge.

\textbf{GCFL+}~\citep{xie2021gcfl} focuses on graph-level FGL by grouping clients based on their GNN gradient trajectories to manage structural and feature diversity. It utilizes dynamic time warping on gradient sequences to ensure the clustering process is stable and robust.

\textbf{FedStar}~\citep{tan2023fgl_fedstar} facilitates graph-level FGL by separating the learning of structures and features. Clients utilize an independent encoder to share domain-invariant structural information while maintaining personalized local encoders for features, thereby minimizing misalignment across the federation.
\ding{184} \textbf{Federated Adaptations of Centralized GFM Approaches.} This category adapts leading centralized graph foundation model (GFM) training paradigms to a federated environment. Specifically, we evaluate Fed-versions of OFA~\citep{gfm_ofa}, GFT~\citep{gfm_gft}, UniGraph~\citep{gfm_unigraph}, GQT~\citep{gfm_gqt}, and GraphCLIP~\citep{gfm_graphclip}. While these methods originally rely on centralized, self-supervised pre-training across a unified dataset, their federated adaptations distribute the pre-training workload across participating clients. 

\textbf{Implementation Details:} During the pre-training phase, each client executes two local optimization epochs using its private data according to the respective framework's objectives. Following this, all trainable weights are transmitted to a central server for parameter averaging. The resulting global model is then synchronized back to all clients to initialize the subsequent round of local training.

\textbf{OFA$^*$}~\citep{gfm_ofa} serves as a benchmark for GFM, designed to extract transferable representations from text-attributed graphs across diverse domains and tasks. It utilizes structured natural language prompts to standardize node and edge descriptions, mapping heterogeneous graph data into a common vector space. Furthermore, it uses nodes-of-interest prompts to unify diverse graph-based tasks within a single predictive framework.

\textbf{GFT$^*$}~\citep{gfm_gft} identifies message-passing computation trees as universal, transferable structural motifs. It utilizes a gVQ-VAE architecture to encode these trees into a discrete codebook. By performing self-supervised reconstruction across multi-domain graphs during pre-training, it develops a foundation model with cross-graph generalization capabilities.

\textbf{UniGraph$^*$}~\citep{gfm_unigraph} facilitates cross-domain transfer by converting heterogeneous graphs, including those without native textual attributes, into a unified linguistic representation. Its cascaded architecture integrates language models and GNNs to jointly capture topological structure and semantic information. The framework adopts Masked Graph Modeling for large-scale self-supervised pre-training and further applies graph instruction tuning with LLMs to enhance zero-shot and few-shot generalization.

\textbf{GQT$^*$}~\citep{gfm_gqt} features a graph quantized tokenizer that separates the tokenization process from the primary Transformer training. By utilizing multi-task self-supervised learning and residual vector quantization, GQT generates hierarchical discrete tokens. This approach reduces the model's memory footprint while simultaneously improving its ability to generalize across graph distributions.

\textbf{GraphCLIP$^*$}~\citep{gfm_graphclip} targets the dependencies on labeled data and the lack of transferability often found in text-attributed graph models. It utilizes LLM-curated graph-summary pairs for self-supervised contrastive pre-training. Through invariant learning, GraphCLIP strengthens zero-shot transferability and introduces a prompt-tuning strategy for few-shot scenarios to prevent the loss of previously acquired knowledge.

\ding{185} \textbf{Federated Graph Foundation Models.}
represents a paradigm shift in decentralized machine learning, designed to bridge the gap between the cross-domain power of foundation models and the privacy-preserving requirements of distributed data. Unlike centralized GFMs, which assume unrestricted access to multi-domain graphs within a single server, FedGFMs~\citep{zhu2025fedgfm} allow independent institutions to collaboratively train a general-purpose encoder without sharing their raw, sensitive graph data. This distinction is critical in real-world scenarios where data is siloed due to institutional constraints, privacy regulations, or competitive interests. By utilizing a federated pre-training pipeline that is typically built on a lightweight graph vector quantization masked auto-encoder backbone (gVQ-MAE), FedGFMs exploit cross-silo computational resources to learn universal topological and semantic patterns while strictly preserving data locality.

\textbf{FedGFM+}~\citep{zhu2025fedgfm} addresses the challenge of knowledge entanglement, a phenomenon where decentralized pre-training on domain-specific graphs causes the global model to produce indistinguishable representations that hinder downstream adaptation. Before pre-training, clients generate domain-specific prototypes that serve as semantic anchors; the server then initializes the global codebook with synthetic embeddings generated through controlled perturbations around these anchors to provide a strong inductive bias for domain separation. During the learning process, each client independently maintains lightweight graph prompts to capture local preferences. These prompts are ultimately consolidated into a shared pool during fine-tuning, allowing the GFM to selectively augment target graph features with relevant domain-specific priors.

\textbf{FedBook}~\citep{fedbook} introduces a systematic server-side aggregation strategy designed to optimize the global codebook by balancing intra-domain coherence with inter-domain diversity. In the first phase, the server identifies semantically similar knowledge units across clients and utilizes a frequency-guided alignment mechanism where low-frequency (potentially unreliable) tokens are refined by referencing high-frequency tokens that serve as stable semantic anchors. The second phase preserves unique domain knowledge by quantifying the domain distinctiveness of each client's local codebook relative to the federation. During the final aggregation, FedBook assigns higher weights to clients with high distinctiveness scores, ensuring that heterogeneous semantics are preserved in the global GFM rather than being diluted by common patterns.

\section{Training Procedure}
\label{appendix: training procedure}
Algorithm~\ref{alg:fedgala} summarizes the end-to-end training procedure of FedGALA, covering the federated pre-training phase and the local prompt-based fine-tuning phase described in the main paper.

\begin{algorithm}[t]
\caption{Training procedure of FedGALA}
\label{alg:fedgala}
\begin{scriptsize}
\begin{tabbing}
\textbf{Input:} \scriptsize Rounds $T, T_{ft}$; Clients $\mathcal{K}$; History $R$; Prompt Pools $\mathcal{P}_T, \mathcal{P}_G$ \\
\textbf{Output:} \scriptsize Pre-trained Encoders $\Theta$; Optimized Indices $m_k^*$; Final Prompts $\Phi$ \\
\textbf{Phase I: Federated Pre-training} \\
1: Initialize $\theta_{\text{str}}, g_{\psi}$; \\
2: \textbf{for} $t = 1$ \textbf{to} $T$ \textbf{do} \\
3: \quad \textbf{Parallel Client Update ($k \in \mathcal{K}$):} \\
4: \quad \quad Download $\theta_{\text{global}}^{(t)}$ and $\{\theta_{\text{his}}^{(r)}\}_{r=1}^R$; \\
5: \quad \quad Compute weights $\boldsymbol{\beta}$ and fuse $\theta_{\text{local}}$ via Eq.~\ref{eq: hist-matching fusion}; \\
6: \quad \quad Update $\theta_{\text{local}}$ and $\theta_{\text{str}}$ by minimizing $\mathcal{L}_{\text{local}}$ via Eq.~\ref{eq: loss function}; \\
7: \quad \quad Upload $\theta_{\text{str}}$ and topology-aware density $\bar{d}_k$ to server; \\
8: \quad \textbf{Server Side Aggregation:} \\
9: \quad \quad Update $\theta_{\text{global}}^{(t+1)}$ via Eq.~\ref{eq: pre-training aggregation} and refresh history pool $\mathcal{H}$; \\
10: \textbf{end for} \\
\textbf{Phase II: Local Prompt-based Fine-tuning} \\
11: Initialize $\mathcal{P}$ with global class-wise Tokens $\mathbf{P}_c$ via Eq.~\ref{eq: class-wise token aggregation}; \\
12: \textbf{for} $t = 1$ \textbf{to} $T_{ft}$ \textbf{do} \\
13: \quad \textbf{Parallel Client Adaptation ($k \in \mathcal{K}$):} \\
14: \quad \quad Identify optimal $m_{k, T}^*, m_{k, G}^*$ via local probing (Eq.~\ref{eq: prompt selection}); \\
15: \quad \quad Freeze $\Theta_{\text{frozen}}$; Update prompts $\phi_{m_k^*}^{\text{type}}$ via Eq.~\ref{eq:2nd phase loss function =}; \\
16: \quad \quad Upload fine-tuned prompts $\phi_{local}$ and selected indices $m_k^*$; \\
17: \quad \textbf{Server Side Specialized Aggregation:} \\
18: \quad \quad Organize clients into latent groups $\mathcal{S}_m$; \\
19: \quad \quad Perform group-aware aggregation for $\mathcal{P}$ via Eq.~\ref{eq: prompt aggregation}; \\
20: \textbf{end for}
\end{tabbing}
\end{scriptsize}
\end{algorithm}

\section{More Experimental Details}

\noindent \textbf{Architecture.} Baseline isolated models utilize two-layer architectures with 64 hidden units, whereas FL/FGL methods adopt GraphSAGE for node/edge tasks and GIN for graph-level tasks. GFM-based frameworks replicate the original reported backbones to ensure consistency. FedGALA employs a 768-dimensional, 2-layer GraphSAGE-based encoder aligned with Sentence-BERT, which remains frozen during prompt-based fine-tuning to ensure stability.

\noindent \textbf{Evaluation Metrics.} To evaluate this representational stability, we report the mean and variance across 10 standardized runs for three clients per dataset. Accuracy and AUC-ROC are utilized for node/edge and graph classification, respectively. Evaluation protocols vary by category: isolated models are assessed locally, FL/FGL after global training, and GFM-based methods via task-specific local fine-tuning.

\noindent \textbf{Environments.} The experimental machine is an Intel(R) Xeon(R) Gold 6240 CPU @ 2.60GHz and NVIDIA A100 with 80GB memory and CUDA 12.4. The operating system is Ubuntu 22.04.5 with 251GB of memory.

\section{Complexity Analysis for FedGFMs}
\label{appendix: complexity analysis}
To ensure a rigorous evaluation, we analyze the computational and communication overhead of \textit{FedGALA} alongside other federated graph foundation models (FedGFMs). We argue that a comparative analysis must be restricted to the FedGFM group to maintain academic fairness; traditional FGL algorithms are designed for task-specific GNNs with shallow features, whereas FedGFMs incorporate high-dimensional Pre-trained Language Models (PLMs) and complex self-supervised objectives. Comparing these disparate scales would result in an unbalanced assessment where the GFM's baseline complexity obscures the actual efficiency of the federated coordination mechanism.

We define $|V|$ and $|E|$ as the number of nodes and edges per client, $d$ as the hidden dimension, $L$ as the layer depth, $K$ as the discrete codebook size, $M$ as the prompt pool size, and $C$ as the number of clients. The computational complexity of a standard GNN message-passing backbone is formally expressed as $O(L(|V|d^2 + |E|d))$.

\textbf{FedGFM+~\citep{zhu2025fedgfm}} involves significant overhead due to its dual-module design for mitigating knowledge entanglement. The client-side complexity is $O(L(|V|d^2 + |E|d) + |V|Kd + |V|Md)$, where the terms $|V|Kd$ and $|V|Md$ represent the costs of nearest-neighbor codebook search and adaptive prompt selection, respectively. Its communication cost is $\Omega(|\Theta| + Md)$, which scales with the size of the prompt pool.

\textbf{FedBook~\citep{fedbook}} utilizes a two-phase aggregation strategy to optimize the global codebook. While its client-side complexity is $O(L(|V|d^2 + |E|d) + |V|Kd)$, the server-side alignment of local codebooks introduces a complexity of $O(C^2 Kd)$, creating a potential bottleneck as the federation grows. The per-round communication payload is $O(|\Theta| + Kd)$.

\textbf{FedGALA (Ours)} demonstrates superior efficiency by leveraging a structural-semantic alignment approach that avoids the irreversible knowledge loss of discrete quantization. By bypassing the codebook search entirely, its client-side complexity is reduced to $O(L(|V|d^2 + |E|d) + |V|d^2)$, making it independent of both the codebook size $K$ and prompt pool size $M$. Crucially, FedGALA optimizes communication to $O(|\Theta|)$, ensuring maximum scalability for large-scale deployments while maintaining the integrity of cross-modal knowledge transfer.

\fi

\end{document}